\documentclass{sig-alternate-05-2015}
\usepackage{times}
\usepackage{url}
\usepackage{latexsym}
\usepackage[ruled]{algorithm2e}
\usepackage{amsmath}
\usepackage{graphicx}

\DeclareMathOperator*{\argmax}{arg\,max}

\usepackage{etoolbox}

\usepackage{tikz}
\usetikzlibrary{calc}

\makeatletter

\patchcmd{\maketitle}{\@copyrightspace}{}{}{}

\makeatother
\begin{document}

\setcopyright{acmcopyright}

\conferenceinfo{CIKM MoST-Rec}{2019, Beijing, China}

\title{Bayesian Optimization for \\Selecting Efficient Machine Learning Models} 

\numberofauthors{3} 

\author{
\alignauthor
Lidan Wang\\
       \affaddr{Adobe Research}\\
       \email{lidwang@adobe.com}
\alignauthor
Franck Dernoncourt\\
       \affaddr{Adobe Research}\\
       \email{dernonco@adobe.com}
\alignauthor
Trung Bui\\
       \affaddr{Adobe Research}\\
       \email{bui@adobe.com}
}

\date{30 July 1999}

\maketitle

\begin{tikzpicture}[remember picture, overlay]
\node at ($(current page.north) + (-0in,-0.5in)$) {Published at CIKM MoST-Rec 2019};
\end{tikzpicture}
\begin{abstract} 

The performance of many machine learning models depends on their hyper-parameter settings. Bayesian Optimization has become a successful tool for hyper-parameter optimization of machine learning algorithms, which aims to identify optimal hyper-parameters during an iterative sequential process. However, most of the Bayesian Optimization algorithms are designed to select models for effectiveness only and ignore the important issue of model training efficiency. Given that both model effectiveness and training time are important for real-world applications, models selected for effectiveness may not meet the strict training time requirements necessary to deploy in a production environment. In this work, we present a unified Bayesian Optimization framework for jointly optimizing models for both prediction effectiveness \textit{and} training efficiency. We propose an objective that captures the tradeoff between these two metrics and demonstrate how we can jointly optimize them in a principled Bayesian Optimization framework. Experiments on model selection for recommendation tasks indicate models selected this way significantly improves model training efficiency while maintaining strong effectiveness as compared to state-of-the-art Bayesian Optimization algorithms. 

\end{abstract}

\section{Introduction}

The performance of many machine learning algorithms depends on their hyper-parameters. For example, the prediction accuracy of support vector machines depends on the kernel and regularization hyper-parameters $\gamma$ and $C$, and deep neural networks are sensitive to a wide range of hyper-parameters, including the number of units per layer, learning rates, weight decay, and dropout rates etc.~\cite{montavon}. It is well-known that hyper-parameter settings often make the difference between mediocre and state-of-the-art performance~\cite{efficientAssessPaper}. As a result, hyper-parameter optimization has been receiving an increasingly amount of attention in the NLP and machine learning communities~\cite{AutoWeka:2013,sklearnPaper,algForHyper:2011,d2015,collaborativePaper,d2016b,lazyPairPaper,d2016}. However, identifying the best model configuration is often a cumbersome process because it can involve several trials and errors before an optimal hyper-parameter setting can be found. Bayesian Optimization~\cite{scalableBayes,speedUp,EfficientBenchmarking:2015,practicalPaper:2012,smacPaper,surrogateBenchmarks} has emerged as an efficient framework for carrying out the model selection process, achieving impressive successes. For example, in several studies, it found better instantiations of convolutional network hyper-parameters than domain experts~\cite{fastBayes}. The common theme is to perform a set of iterative hyper-parameter optimizations. In each round, these methods fit a hyper-parameter response surface using a probabilistic regression function such as Gaussian Process~\cite{practicalPaper:2012} or tree-based models~\cite{smacPaper}, where the response surface maps each hyper-parameter setting to an approximated accuracy. The learned regression model is then used as a surrogate of the response surface to explore the search space and identify promising hyper-parameter candidates to evaluate next in order to enhance validation accuracy.

While these methods have enjoyed great success compared to conventional random search~\cite{randomSearchPaper,randomSearchPaper2} and grid search algorithms for model selection, the focus of these work have largely been on optimizing for \textit{effectiveness}, while ignoring the resulting model's training efficiency. Given that both prediction accuracy and model training time are important for real-world applications, models selected for effectiveness may not meet the strict real-world efficiency requirements necessary to deploy in a production environment. In addition, most of the previous methods exclusively focus on optimizing the hyper-parameters of a given model class, while ignoring other important extrinsic hyper-parameters such as training set size which can influence both speed and accuracy. For example, model training time typically grows proportionally with respect to training set size, and the prediction accuracy can also be influenced by the amount of training data used for learning. If the tolerance for inefficient model training is low, then the amount of training data should be reduced or adjusted with the rest of intrinsic hyper-parameters to meet the stringent efficiency requirements.

Given that both model effectiveness and training time are important for real-world applications, in this work, we propose a unified Bayesian Optimization framework for jointly selecting models for prediction effectiveness \textit{and} training efficiency. First, we propose an objective that captures the tradeoff between these two metrics, then we demonstrate how we can jointly optimize them in a principled Bayesian Optimization framework. In addition, we account for extrinsic hyper-parameters such as training set size in the hyper-parameter optimization space. We will demonstrate this joint optimization of both measures in an enriched hyper-parameter space leads to selecting more efficient and accurate models. It is important to point out our work is fundamentally different from previous Bayesian Optimization that considers the speed of the hyper-parameter search/model selection process~\cite{fastBayes,scalableBayes,practicalPaper:2012,nlpEfficiency} -- our focus is on model training efficiency, in addition to accuracy (as part of the joint optimization), while their focus is on hyper-parameter search efficiency. Our work can be viewed as taking an efficiency-centric view at selecting effective models. Experiments on model selection for recommendation and question answering tasks indicate models selected this way significantly improves model training efficiency while maintaining strong effectiveness as compared to state-of-the-art Bayesian Optimization algorithms.

The remainder of the paper is organized as follows: We start with a discussion of related work. Next, in Section~\ref{method} we propose metrics for quantifying the tradeoff between prediction accuracy and training efficiency, then discuss methods for model selection based on the tradeoff metric. Section~\ref{experiments} presents experimental results under different tradeoff scenarios for recommendation and question answering tasks, before concluding in Section~\ref{conclusion}.

\section {Related Work} 

Model selection is an important problem in machine learning applications. However, it is typically a time consuming process if the hyper-parameter space and training set size are large. Traditional grid-search can not easily scale to a large number of hyper-parameters efficiently. In this setting, several Bayesian Optimization algorithms have been proposed to speed up model selection~\cite{scalableBayes,speedUp,EfficientBenchmarking:2015,tutorial,smacPaper,surrogateBenchmarks}. They typically work by iteratively fitting a surrogate regression function to estimate the performance of different settings of hyper-parameter values, and based on the estimation, certain hyper-parameter candidate values are evaluated, which in turns improves the regression function, and helps with better identification of top hyper-parameter values in the next round. However, their focus is on finding good hyper-parameters and models for prediction accuracy, rather than jointly considering both accuracy and training efficiency. 

Several work has focused on the efficiency of the model selection process~\cite{fastBayes,scalableBayes,autoKeras,multtaskPaper:2013,nlpEfficiency}. Their main goal is to improve the speed of hyper-parameter tuning. As mentioned earlier, their work is different from the focus of this work --  our focus is on selecting models for both model training efficiency and accuracy. In a production setting, model training time is an important metric for evaluating the practical usefulness of a given model. Thus, we aim to provide a principled way for selecting models that achieve an optimal balance between prediction accuracy and training time.

Furthermore, additional approaches have been proposed for speeding up Bayesian Optimization process for model selection by using search time as an additional constraint~\cite{speedConstrained,speedConstrained2,speedConstrained3,practicalPaper:2012}. Similar to above, these work either focuses on the speed of model selection, or cannot control the tradeoff between accuracy and training time in a tunable way.

We note our approach is orthogonal and complementary to parallel Bayesian Optimization~\cite{practicalPaper:2012} and multi-task learning~\cite{efficientTransferPaper:2014,multtaskPaper:2013}, because the improved model training efficiency, as achieved by our algorithm, is a basic building block of the other algorithms, thus can directly help the efficiency of multiple parallel runs~\cite{practicalPaper:2012}, as well as runs across different datasets~\cite{efficientTransferPaper:2014,multtaskPaper:2013}.

\section{Method} \label{method}

Most of the previous Bayesian Optimization algorithms focus on selecting models for their accuracy. However, from a practical point of view, we are also concerned with model training time. For example, there is an inherent tradeoff between model training efficiency and prediction accuracy. While complex deep neural networks preform very well on a variety of natural language tasks, they typically take longer time to train than networks with simpler structures and fewer number of hyper-parameters. We would like to carefully trade off these two competing measures and obtain a model that can simultaneously be effective and efficient (training time).

In addition, building highly efficient models given the many real-world constraints is desirable from several points of view. First, real-world machine learning systems are diverse and have different tolerances to training latency. Some need results as soon as possible, even if the results may be of slightly lower quality, while for others, waiting a bit longer may be acceptable if it means a better model. Furthermore, how a system behaves under load depends on the exact architecture, but accuracy-efficiency tradeoff can help the system adapt. For example, when the number of training jobs is high, we might tighten the tradeoff parameter to focus more on efficiency in order to produce results quickly -- this is more preferable than forcing users to wait longer for results. Before discussing the optimization algorithm, we first describe the tradeoff objective used by our Bayesian Optimization algorithm.

\subsection {Measuring Training Efficiency}

Our goal is to automatically select models that better adapt to time requirements by optimally trading off between prediction accuracy and training efficiency. Therefore, we need to capture model training efficiency in the optimization objective in addition to accuracy.

There are many ways to measure the training efficiency $\sigma(\boldsymbol{\lambda})$ of a given model. The most straightforward way is the wall-clock time. It needs to be normalized into the range of $[0, 1]$ in order to be used as an optimization metric. This framework takes the normalized efficiency as an input to the optimization framework. In addition to wall-clock time, it is also possible to create analytical models of training time. However, depending on complexity of the machine learning model, current analytical formula may or may not be able to cover all types of models sufficiently well. It remains our future work to explore a general analytical technique for this direction.

\subsection {Measuring Prediction Accuracy}

There has been a great deal of research into evaluating the effectiveness of systems. Therefore, we simply make use of existing effectiveness measures here. We define the effectiveness of a model configuration $\boldsymbol{\lambda}$ as $L(\boldsymbol{\lambda})$. As with the efficiency metrics, we are primarily interested in effectiveness measures with range $[0, 1]$. Most of the commonly-used effectiveness metrics satisfy this property, including accuracy, precision, and recall. In this paper we will exclusively focus on accuracy as the effectiveness metric of interest, although any of the above metrics can be substituted in the optimization framework without loss of generality.

\subsection{Accuracy-Efficiency Tradeoff Metric} \label{tradeoffMetric}

Our goal is to automatically select models that can be both effective and efficient. Our objective function, which we call Accuracy-Efficiency tradeoff $T$, is defined as a weighted linear combination of accuracy and training efficiency:

$$T_\alpha (\boldsymbol{\lambda}) =  L(\boldsymbol{\lambda}) - \alpha \cdot \sigma(\boldsymbol{\lambda})$$

where $\alpha \ge 0 $ is a parameter that controls the relative importance between effectiveness and efficiency. We chose this formulation so that the case $\alpha = 0$ corresponds to the standard accuracy objective used in model selection.

Note that $T$ is not a single measure, but a family of measures, parameterized by the tradeoff parameter $\alpha$. By varying $\alpha$ we can capture a full spectrum of accuracy/efficiency tradeoffs: from $\alpha$ = 0 (standard default setting) to larger values of $\alpha$ that lead to more efficient models, or small values of $\alpha$ that lead to effectiveness-centric models.

\subsection {Bayesian Optimization Algorithm}

In this work, we design a Bayesian Optimization strategy that jointly selects models for both accuracy and model training efficiency, by optimizing the tradeoff metric. Furthermore, our method leverages the training set size as another hyper-parameter optimized by the procedure, in addition to a given model's standard hyper-parameters. We evaluate our tasks on recommendation and related problems. This work is different from previous methods both in terms of what we optimize and our focus on achieving a tunable balance between the two competing measures, although it can also have a positive impact on the optimization procedure's speed as a side effect.

\subsubsection {Enriching Hyper-parameter Space for Efficiency}

Traditional Bayesian hyper-parameter optimizations exclusively focus on optimizing the hyper-parameters of a given model class, while ignoring other important extrinsic hyper-parameters such as training set size which can influence both speed and accuracy. For example, as dataset size grows, even simple models such as logistic regression can require more training time, leading to increased overall time cost. Thus, training set size is an important factor that controls training efficiency and should be accounted for during the optimization process. In this work, we leverage dataset size as an additional degree of freedom enriching the representation of the optimization problem, along with model's hyper-parameters. Similar to other hyper-parameters, training set size can be varied (from original set size to a fraction of the original size) to better guarantee the accuracy-efficiency tradeoff. Its impact will be jointly considered with other hyper-parameters, and the selected training set size will be used for training a final model for evaluation purpose.

Let $\boldsymbol\lambda=\{\lambda_1, \ldots, \lambda_m\}$ denote the enriched hyper-parameters of a machine learning algorithm, and let $\{\Lambda_1, \ldots, \Lambda_m\}$ denote their respective domains. When trained with $\boldsymbol\lambda$ on training data, the validation accuracy is denoted as $L(\boldsymbol{\lambda})$. The goal of hyper-parameter optimization is to find a hyper-parameter setting $\boldsymbol\lambda^*$ such that accuracy is maximized. Current state-of-the-art methods have focused on using model-based Bayesian Optimization~\cite{practicalPaper:2012,smacPaper} to solve this problem due to its ability to identify good solutions within a small number of iterations as compared to more conventional methods.

\subsubsection {Bayesian Optimization for Hyper-parameter \mbox{ } Learning}\label{sec:bayesOpt}

Model-based Bayesian Optimization~\cite{tutorial} starts with an initial set of hyper-parameter settings $\boldsymbol\lambda_1, \ldots \boldsymbol\lambda_k$, where each setting denotes a set of assignments to all hyper-parameters. These initial settings are then evaluated on the validation data and their accuracies are recorded. The algorithm then proceeds in rounds to iteratively fit a probabilistic regression model $V_L$ to the recorded accuracies. A new hyper-parameter configuration is then suggested by the regression model $V_L$ with the help of an acquisition function~\cite{tutorial}. Then the accuracy of the new setting is evaluated on validation data, which leads to the next iteration. A common acquisition function is the expected improvement over accuracy, EI~\cite{tutorial}, over best validation accuracy seen so far $L^*$:

\begin{eqnarray} \label{eqn_acq_acc}
a_L(\boldsymbol{\lambda}, V_L) = \int_{-\infty}^{\infty} max (L-L^*, 0) p_{V_L}(L|\boldsymbol{\lambda}) d_{L} \\ \nonumber
\end{eqnarray}

where $p_{V_L}(L|\boldsymbol\lambda)$ denotes the probability of accuracy $L$ given configuration $\boldsymbol \lambda$, which is encoded by the probabilistic regression model $V_L$. The acquisition function is used to identify the next candidate (the one with the highest expected improvement over current best $L^*$). More details of acquisition functions can be found in~\cite{tutorial}.

The most common probabilistic regression model $V_L$ is the Gaussian Process prior~\cite{practicalPaper:2012}, which is a convenient and powerful prior distribution on functions. For the purpose of our experiments, we also use Gaussian Process prior as the regression model. However, we would like to note the fact that the proposed optimization is agnostic of the regression model used, and can easily handle other instantiations of the regression model.

\subsubsection {Optimizing the Tradeoff Metric}

The acquisition function used in standard Bayesian Optimization is for accuracy only, and not directly applicable to the tradeoff metric. Just as we do not know the true objective function $L(\boldsymbol{\lambda})$ for accuracy, we also do not know the training duration function $\sigma(\boldsymbol{\lambda})$. We can nevertheless employ our Gaussian process machinery to model $\sigma(\boldsymbol{\lambda})$ alongside $L(\boldsymbol{\lambda})$. 

In this work, we assume that these functions are independent of each other. Under the independence assumption, the acquisition function for the tradeoff metric can be written as follows:

\begin{eqnarray} \label{eqn_acq_tradeoff}
a_T(\boldsymbol{\lambda}) = a_L(\boldsymbol{\lambda}, V_L) - \alpha \cdot a_{\sigma}(\boldsymbol{\lambda}, V_{\sigma}) \\ \nonumber
\end{eqnarray}

where $a_L(\cdot)$ is the acquisition function for accuracy-based selection, as defined by Equation~\ref{eqn_acq_acc} above, and $a_{\sigma}(\cdot)$ is the acquisition function for efficiency-based selection which is similarly defined as:

\begin{eqnarray} \label{eqn_acq_efficiency}
a_{\sigma}(\boldsymbol{\lambda}, V_{\sigma}) = \int_{-\infty}^{\infty} max ({\sigma}-{\sigma}^*, 0) p_{V_{\sigma}}(\sigma|\boldsymbol{\lambda}) d_{\sigma} \\ \nonumber
\end{eqnarray}

Thus, the acquisition function for the tradeoff metric is a weighted linear combination of the acquisition functions defined on accuracy and training time, and the weight $\alpha$ is the same parameter used to control their relative importance, as defined in Section~\ref{tradeoffMetric}. In addition, the regression function for training time is $V_{\sigma}$, and similarly to accuracy's regression function $V_L$, it is iteratively fitted and improved during the optimization process.

This acquisition function is then used to explore the enriched hyper-parameter search space and to identify promising candidate points to evaluate. More specifically, in each Bayesian iteration, the hyper-parameter setting predicted to yield most information about the tradeoff metric, as given by the acquisition function will be evaluated.

\begin{algorithm}[t]
    \label{algo:alg}
    \caption{Bayesian Optimization for Learning based on Tradeoff Metric}
    \KwIn{Loss function $T$, number of iterations $S$, initialization $\boldsymbol\lambda_{1:k}$}
    \KwOut{Selected model $\boldsymbol\lambda^*$}
    \For{i=1 to $k$} {
       $L_i$ = Evaluate L($\boldsymbol\lambda_i$) \\
       $\sigma_i$ = Evaluate $\sigma(\boldsymbol\lambda_i$) \\
    }
    \For{j=k+1 to $S$} {
       $V_L$: regression model on $\big \langle \boldsymbol\lambda_i, L_i \big \rangle_{i=1}^{j-1}$ \\
       $V_\sigma$: regression model on $\big \langle \boldsymbol\lambda_i, \sigma_i \big \rangle_{i=1}^{j-1}$ \\
       $\boldsymbol\lambda_j =   \argmax_{\boldsymbol \lambda \in \boldsymbol \Lambda} \displaystyle a_T(\boldsymbol \lambda)$ \\
       $L_j$ = Evaluate L($\boldsymbol\lambda_j$) \\
       $\sigma_j$ = Evaluate $\sigma(\boldsymbol\lambda_j$) \\
    }
    \textbf{return} {$\boldsymbol \lambda^*= \argmax_{\boldsymbol \lambda_\in  \{\boldsymbol \lambda_1, \ldots, \boldsymbol \lambda_S \}} L(\boldsymbol \lambda) - \alpha \cdot \sigma (\boldsymbol \lambda)$}

\end{algorithm}

\begin{table}
\centering
\begin{tabular}{|p{2cm}|p{4cm}|}  
\hline & \bf Enriched space of hyper-parameters \\ \hline
Deep Learning LSTM & training set size, dropout, regularization parameter, hidden units, batch size, num epochs \\  \hline
SVM  & training set size, bias, cost parameter, regularization parameter \\   
\hline
\end{tabular}
\caption{Enriched hyper-parameter space used in models (training set size is an additional hyper-parameter for tuning).}
\label{hyperparam_list} 
\vspace{-0.3cm}
\end{table}

Our model selection algorithm for tradeoff metric is shown in Algorithm~\ref{algo:alg}. It can be viewed as a generalization of the standard Bayesian optimization algorithm (i.e., when $\alpha$=0). It proceeds in a number of rounds. It starts by running training with a set of initial hyper-parameter values\footnote{We adopt the convention that a Sobol sequence is used to initialize the first stage~\cite{practicalPaper:2012}. The value $k$ for the first stage is the number of points in the Sobol sequence.} in an initial design and recording the resulting validation accuracies ($L_i$'s), and training efficiencies ($\sigma_i$'s) from these settings. Afterwards, it iterates the following three stages: (1) fit Gaussian probabilistic regression functions $V_L$ for accuracy, and $V_\sigma$ for training efficiency over the pairs of collected hyper-parameter settings and the collected accuracies/time values; (2) use the probabilistic models $V_L$ and $V_\sigma$ to compute tradeoff acquisition function $a_T(\boldsymbol{\lambda})$ (Equation~\ref{eqn_acq_tradeoff}), which is then used to select a promising input $\lambda_j$ to evaluate next by quantifying the desirability of obtaining the function value at $\lambda_j$ through $a_T(\boldsymbol{\lambda})$. Then we run model training with the selected hyper-parameter candidate and record the accuracy ($L_j$) and efficiency ($\sigma_j$) for regression model in the next round. Note the hyper-parameter space we search through is the enriched hyper-parameter space which augments the original space with training set size, with candidates selected from varying fractions of full size $\{20\%, 40\%, 60\%, 80\%, 100\%\}$.

As mentioned earlier, this optimization algorithm subsumes the standard Bayesian optimization algorithm as a special case when tradeoff parameter = 0 (only accuracy is considered). For the number of top configurations $k$ used to initialize the first stage, we know the larger $k$ is, the better results in the next stage since Bayesian Optimization relies on good initial knowledge to fit good regression models~\cite{initializingPaper:2015}. However, larger $k$ value also leads to high computation cost, since these initial settings will have to be evaluated first. In practice, the number of iterations $T$ and the value of $k$ depend on the quantity of the data and the quality of machine learning model be selected. In our experiments, we empirically choose their values to be $T=20$ and $k=3$ which result in a good balance between accuracy and speed on the given datasets.

\section{Experiment} \label{experiments}

\begin{figure}[t]
\centerline{\includegraphics[width=2.8in]{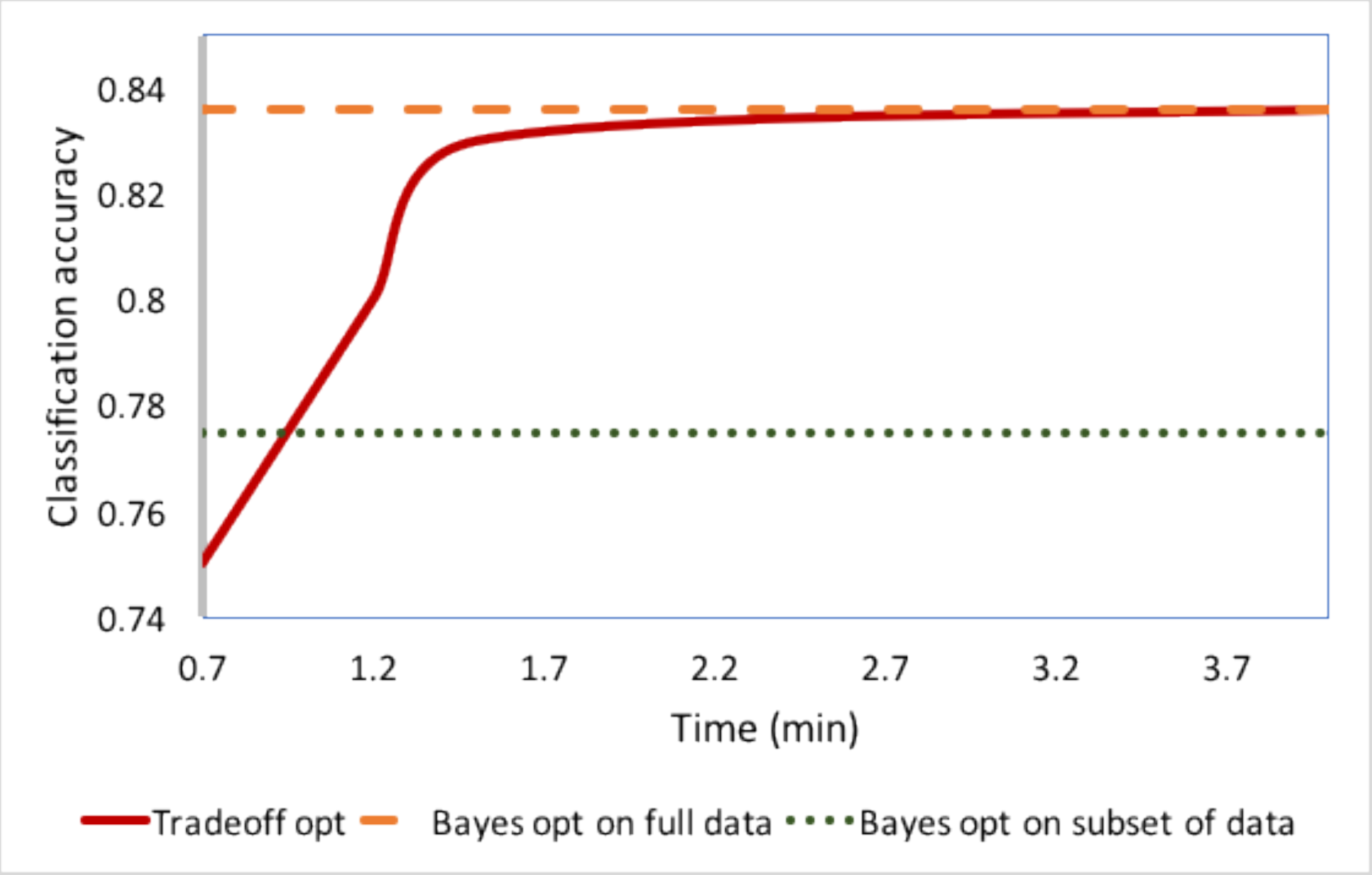}}
\caption{\small Yelp rating prediction: effectiveness vs training time, generated by varying tradeoff parameter $\alpha \in \{0.1, 0.3, 0.5, 0.7, 0.9\}$} 
\label{yelp}
\end{figure}

\begin{figure}[t]
\centerline{\includegraphics[width=2.8in]{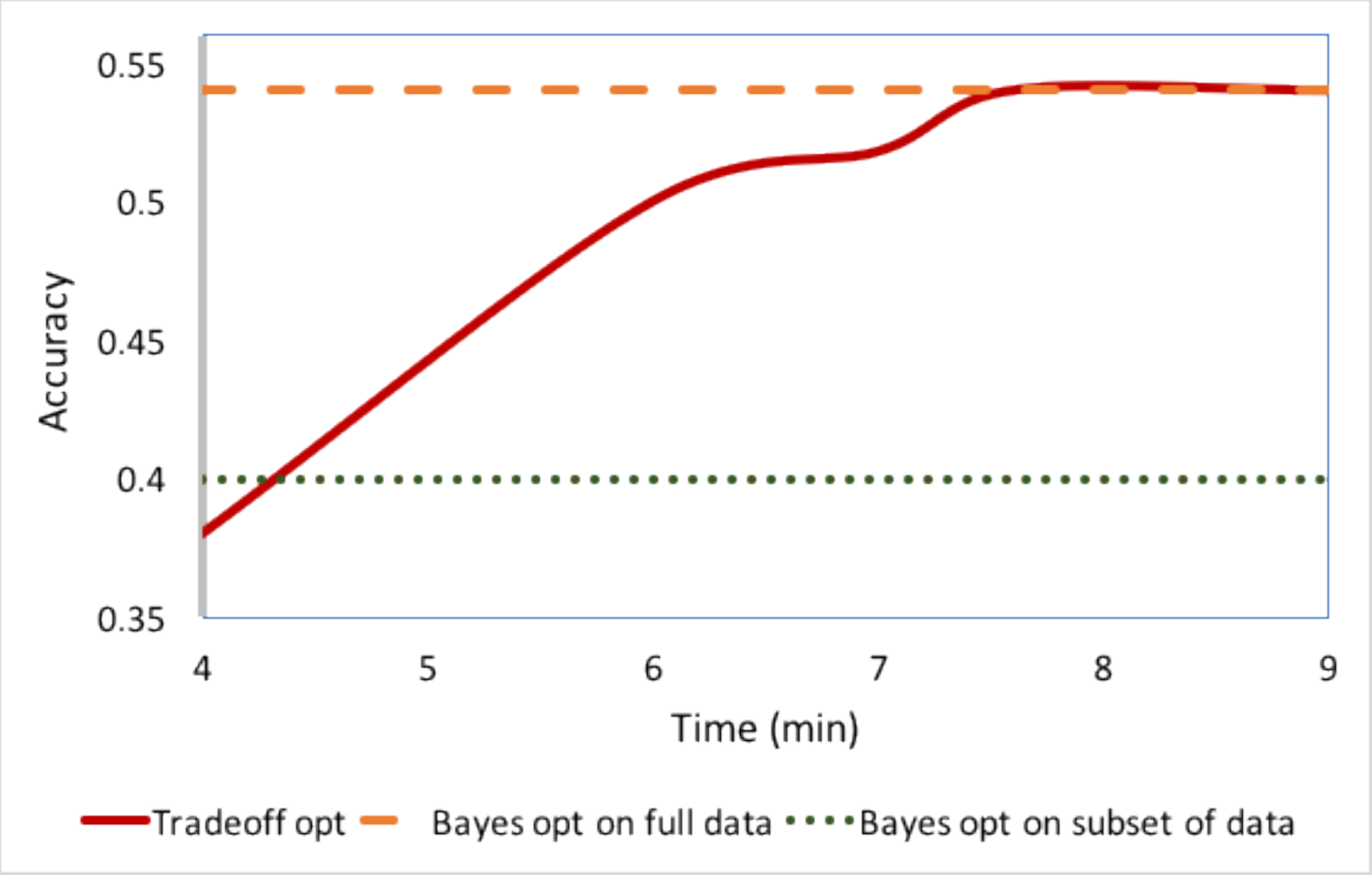}}
\caption{\small Insurance: effectiveness vs training time, generated by varying tradeoff parameter $\alpha \in \{0.1, 0.3, 0.5, 0.7, 0.9\}$} 
\label{homedepot}
\end{figure}

\begin{table*}
\centering
\begin{tabular}{|c|c|c|c|}   
\hline & \bf Insurance (accuracy) & \bf Yelp (accuray) \\ \hline
Bayes opt on full data & 0.54    & 0.84   \\
Bayes opt on small subset & 0.38& 0.78 \\ 
New algorithm & 0.51    & 0.81  \\
\hline
\end{tabular}
\caption{Average test accuracy across time constraints.}
\label{table:avgAccuracy} 
\end{table*}

\begin{table*}
\centering
\begin{tabular}{|c|c|c|c|}  
\hline & \bf Insurance (min) & \bf Yelp (min) \\ \hline
Bayes opt on full data & 12   & 4.0   \\
Bayes opt on small subset & 5.2 & 1.9  \\ 
New algorithm & 6.7    & 2.3  \\
\hline
\end{tabular}
\caption{Average time across time constraints.}
\label{table:avgTimes} 
\vspace{-0.3cm}
\end{table*}

We empirically evaluate the algorithm on the following tasks: product review rating prediction, and question answering.

For product rating prediction, we use the Yelp dataset~\cite{yelpDataset} which is a customer review dataset. Each review contains a star/rating (1-5) for a business, and the task is to predict the rating based on the textual information in the review. The training data contains feature vectors, and unique unigrams are used as features (after standard stop-word removal and stemming~\cite{manningBook}). For question answering, we use a standard benchmark dataset from the insurance domain\footnote{https://github.com/shuzi/insuranceQA}. All datasets contain independent training, validation, and test data, from which the machine learning models are built, accuracies are evaluated, and test results are reported, respectively. 

We evaluate our algorithm method against two methods: 1) state-of-the-art Bayesian Optimization for hyper-parameter learning~\cite{practicalPaper:2012}); and 2) the same Bayesian Optimization but only applied on a small subset of data for training speed. We extend the standard Bayesian Optimization method~\cite{practicalPaper:2012} to multi-objective Bayesian Optimization studied in this paper. We consider selecting models for two machine learning algorithms: SVM implementation for Yelp rating prediction~\cite{liblinear}, which is a standard technique widely applied for this task for balancing speed and performance; deep neural networks for the insurance domain, which consists of a set of BiLSTM models for capturing user search terms and answer candidates, respectively, and computing their similarity score~\cite{lstm}. The details of these algorithms including the enriched hyper-parameter space are shown in Table~\ref{hyperparam_list}. As mentioned earlier, for training set size, the candidates are fractions of the full size at $\{20\%, 40\%, 60\%, 80\%, \\
 100\%\}$.

\subsection {Accuracy vs time}

Figures 1-2 compare the test accuracy of our proposed technique against the two baselines as a function of training time (i.e., models obtained by tradeoff optimization with varying tradeoff parameter values). The state-of-the-art Bayesian optimization is applied on full training data, and the fast variant of Bayesian Optimization is applied with 30\% of training data (randomly sampled from full dataset). The accuracy on test data is reported on the y-axis, and the time is reported on the x-axis.

We note both baselines are not able to adapt to diverse time constraints, and they represent standard fixed point evaluation of Bayesian Optimization algorithm. In contrast, our approach is able to adapt to different tradeoff criteria (obtained by varying tradeoff parameter $\alpha$). While in general, when given more time, the new approach producing models that consistently have higher test accuracy, it is able to reach the upper-baseline result (Bayes optimization on full data) within shorter period of time. For example, for Insurance, it reaches the upper-baseline at training time $7.6$ min (Figure 2), while the full Bayesian Opt takes $12$ min to reach the same accuracy as shown in Table 3 (note the training efficiency of upper-baseline is equivalent to tradeoff $\alpha=0$). Similarly, for the Yelp, the new algorithm is able to reach upper-baseline performance with less time than the standard approach that does not consider training efficiency. The new approach is able to reach the upper-baseline with less time as a result of more efficiently exploring the hyper-parameter space and enriching the hyper-parameter space with efficiency-oriented hyper-parameter (training set size), which helps to provide an extra degree of freedom for exploring the tradeoff metric.

\subsection {Expected Accuracy Across Tradeoff Parameter}

To investigate the average accuracy achieved by different methods across different time points (generated by varying the tradeoff parameter), we compare their mean expected accuracy in Table~\ref{table:avgAccuracy}, and similarly their average speed in Table~\ref{table:avgTimes}. In terms of average accuracy, Bayesian optimization on full training set represents an upper-baseline (since its training time is unconstrained). As expected, it has the highest accuracy, however, it has the slowest model training time as well. For example, for Insurance dataset, while the upper-baseline achieves good accuracy, its training time is significantly slower than the other two methods. Same patterns are observed for Yelp as well. This raises the question whether it is worthwhile to have a highly expensive model for a moderate gain in test accuracy -- especially in a time-sensitive or resource-constrained setting, where time is critical. This balance is exactly captured by the tradeoff metric. By varying the tradeoff parameter, we are able to capture a variety of real-world preferences on accuracy vs training time costs. The new algorithm has similar average accuracy as the upper-baseline, while significantly more efficient at the same time.  For example, it provides $94\%$ of the upper-baseline accuracy under $55\%$ of upper-baseline's time for Insurance, and provides $96\%$ the upper-baseline accuracy under $57\%$ of upper-baseline's time for Yelp as shown in Table 2 and Table 3.

\subsection {Discussions}  

There are other ways to quantify the interaction between training efficiency and accuracy. One example is improvement per second, which is the improvement in accuracy as a function of model training time. However, different from our proposed tradeoff metric, it cannot explicitly provide a tunable tradeoff that controls the balance between the two competing measures. As mentioned before, both real-world applications and users are diverse and have different tolerances for latency. Some may prefer faster results, while others may be willing to wait longer. Thus, it is desirable to have a tunable tradeoff metric that can capture a variety of preferences between the two measures. Our proposed framework directly optimizes the tradeoff for desired results.

\section{Conclusion} \label{conclusion}

We introduced a unified Bayesian Optimization framework for jointly optimizing models for both effectiveness and training efficiency. We propose an objective that captures the tradeoff between accuracy and training efficiency and demonstrate how we can jointly optimize these measures in a principled framework. Experiments on several real-world model selection and rating prediction tasks indicate this approach significantly improves model training efficiency while maintaining strong effectiveness as compared to state-of-the-art baseline models.

\bibliographystyle{abbrv}

\end{document}